\documentclass[10pt,twocolumn,letterpaper]{article}

\usepackage{cvpr}
\usepackage{times}
\usepackage{epsfig}
\usepackage{graphicx}
\usepackage{amsmath}
\usepackage{amssymb}
\usepackage{subcaption}
\usepackage{booktabs}
\usepackage{breqn}
\usepackage[bottom]{footmisc}

\usepackage[pagebackref=true,breaklinks=true,colorlinks,bookmarks=false]{hyperref}
\cvprfinalcopy 

\def\cvprPaperID{934} 

\ifcvprfinal\fi
\begin{document}

\title{Attending to Discriminative Certainty for Domain Adaptation}

\author{{Vinod Kumar Kurmi}\thanks{Equal contributions from both authors.}, \quad  {Shanu Kumar}\footnotemark[1],  \quad  Vinay P. Namboodiri\\
Indian Institute of Technology Kanpur\\
India\\
{\tt\small \{vinodkk,sshanu,vinaypn\}@iitk.ac.in}}

\maketitle

\begin{abstract}
  In this paper, we aim to solve for unsupervised domain adaptation of classifiers where we have access to label information for the source domain while these are not available for a target domain. While various methods have been proposed for solving these including adversarial discriminator based methods, most approaches have focused on the entire image based domain adaptation. In an image, there would be regions that can be adapted better, for instance, the foreground object may be similar in nature. To obtain such regions, we propose methods that consider the probabilistic certainty estimate of various regions and specify focus on these during classification for adaptation. We observe that just by incorporating the probabilistic certainty of the discriminator while training the classifier, we are able to obtain state of the art results on various datasets as compared against all the recent methods. We provide a thorough empirical analysis of the method by providing ablation analysis, statistical significance test, and visualization of the attention maps and t-SNE embeddings. These evaluations convincingly demonstrate the effectiveness of the proposed approach.
\end{abstract}

\section{Introduction}
With the advent of deep learning, there has been substantial progress for solving image classification tasks with state of the art methods obtaining lesser than 3\% error (on top five results) on the imagenet dataset. However, it was observed that these results do not transfer to other datasets~\cite{long_ICML2015} due to the effect of dataset bias~\cite{torralba_CVPR2011}. The classifiers trained on one dataset (termed source dataset) show a significant drop in accuracy when tested on another dataset (termed target dataset). To address this issue, some methods have been proposed for adapting domains~\cite{csurka2017domain,wang2018deep}. One of the more successful approaches towards addressing this domain shift has been based on the adversarial adaptation of features using an adversarial discriminator~\cite{ganin_ICML2015} that aims to distinguish between samples drawn from the source and target datasets. Due to the adversarial training, the feature representations are brought close such that the discriminator is not able to distinguish between samples from source and target dataset. However, all approaches that are based on this idea consider the whole image as being adapted. This usually is not the case as there are predominant regions in an image that may be better adapted and useful for improving classification on target dataset. We address this issue and propose a simple approach for solving this problem.
 
\begin{figure}[!]
\begin{subfigure}{.155\textwidth}
  \centering
  \includegraphics[width=0.9\linewidth]{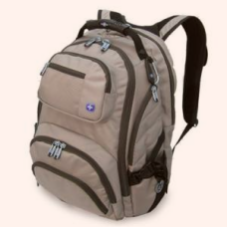}
  \caption{}
  \label{fig:sfig1}
\end{subfigure}
\begin{subfigure}{.155\textwidth}
  \centering
  \includegraphics[width=0.9\linewidth]{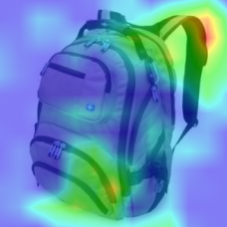}
  \caption{}
  \label{fig:sfig2}
\end{subfigure}
\begin{subfigure}{.155\textwidth}
  \centering
  \includegraphics[width=0.9\linewidth]{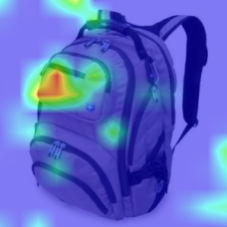}
  \caption{}
  \label{fig:sfig33}
\end{subfigure}
\begin{subfigure}{.155\textwidth}
  \centering
  \includegraphics[width=0.9\linewidth]{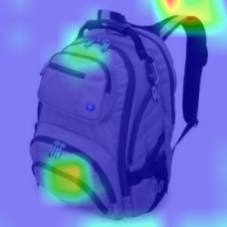}
  \caption{}
  \label{fig:sfig4}
\end{subfigure}
\begin{subfigure}{.155\textwidth}
  \centering
  \includegraphics[width=0.9\linewidth]{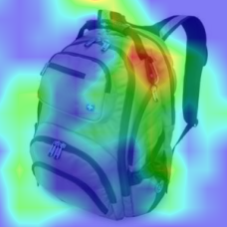}
  \caption{}
  \label{fig:sfig11}
\end{subfigure}
\begin{subfigure}{.155\textwidth}
  \centering
  \includegraphics[width=0.9\linewidth]{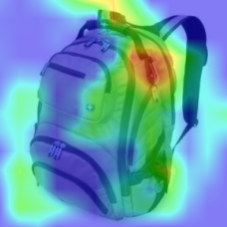}
  \caption{}
  \label{fig:sfig111}
\end{subfigure}
\caption{Visualization of the uncertainty and certainty maps of the discriminator during the midst of training is provided for the input image (\textbf{a}). The aleatoric and predictive uncertain regions of the discriminator are shown in image (\textbf{b}) and (\textbf{d}). While aleatoric and predictive certain regions of the discriminator are shown in (\textbf{c}) and (\textbf{e}). From the figure, it is clear that the certain regions of the discriminator during training mostly corresponds to (\textbf{f}) classifier's activation map based on true label at the end of training.} 
\label{fig:intro}
\vspace{-1.5em}
\end{figure}

 To specify regions that can be adapted we propose the use of certainty of a probabilistic discriminator. During training, we identify regions where the discriminator is certain, i.e., the probabilistic uncertainty for these regions is low. These regions can be adapted because there exists a clear distinction between the source and target regions. Figure~\ref{fig:intro} shows that using measures such as data uncertainty (known as aleatoric uncertainty)~\cite{kendall2017uncertainties} and predictive uncertainty~\cite{lakshminarayanan_nips2017}, we can obtain regions that can be adapted better. We also observe from the Figure~\ref{fig:intro} and ~\ref{fig:dis_train} that for most of the duration during training, discriminator is certain on the foreground regions, as the foreground is hard to adapt. Hence, when the classifier is trained with the emphasis being placed on these regions, then we observe that the classifier focuses on these regions during prediction and therefore generalizes better on target dataset. Quantitatively we observe results that are up to 7.4\% better than the current state of the art methods on Office-Home dataset~\cite{venkateswara_cvpr2017deep}. 

To summarize, through this paper we make the following contributions:
\begin{itemize}
    \vspace{-0.7em}
    \item We propose a method to identify adaptable regions using the certainty estimate of the discriminator, and this is evaluated using various certainty estimates.
    \vspace{-0.7em}
    \item We use these certainty estimate weights for improving the classifier performance on target dataset by focusing the training of the classifier on the adaptable regions of the source dataset.
    \vspace{-0.7em}
    \item We provide a thorough evaluation of the method by considering detailed comparison on standard benchmark datasets against the state of the art methods and also provide an empirical analysis using statistical significance analysis, visualization and ablation analysis of the proposed method.
    \vspace{-0.7em}
    \item An additional observation is that by using Bayesian classifiers we also improve the robustness of the classifier in addition to obtaining certainties of classification accuracy. This aids in better understanding of the proposed method.
\end{itemize}

\section{Literature Survey}

Some studies have examined different adaptation methods. One study by~\cite{tzeng_arxiv2014} examined domain adaptation by minimizing the maximum mean discrepancy distance. The maximum mean discrepancy based approaches were further extended to multi-kernel MMD in~\cite{long_ICML2015}. In adversarial learning framework~\cite{ganin_ICML2015} has proposed a method to minimize the source target discrepancy using a gradient reversal layer at discriminator. Recently many adversarial methods have been applied in the domain adaptation task to bring the source and target distribution closer. Adversarial discriminative domain adaptation~\cite{tzeng_CVPR2017} considers an inverted label GAN loss. Wasserstein distance based discriminator was used in~\cite{shen_AAAI2018wasserstein} to bring the two distributions closer. Domain confusion network~\cite{tzeng_ICCV2015} was also used to solve the adaptation problem in two domains by minimizing the discriminate distance between two domains. The discriminative feedback of the discriminator also applied in the paraphrase generation problem~\cite{patro2018learning}. Another adversarial discriminator based model is~\cite{pei_arxiv2018}, where multiple discriminators (MADA) have been used to solve the mode collapse problem in the domain adaptation. Some works closely related to MADA have been proposed in~\cite{cao2018partial, Cao_2018_CVPR}. The labeled discriminator~\cite{kurmi1904.01341} used to tackle the mode collapse problem in domain adaptation. The adversarial domain adaptation also explored in scene graph~\cite{kumar2019adversarial}. Other source and target discrepancy minimization based methods such as~\cite{saito_cvpr2017maximum,zhang_cvpr2018aligning} also address the domain adaptation problem. \cite{patro2018multimodal,Patro_2018_CVPR} have proposed an exemplar based discrepancy minimization method. Recently~\cite{bousmalis_CVPR2017,choi2017_cvprstargan} have applied generative adversarial network~\cite{goodfellow_NIPS2014} for the domain adaptation problems. Image generation methods are used to adapt the source and target domain~\cite{ghifary_ICCV2015,sankaranarayanan_cvpr2018learning,murez_cvpr2018image}. Other work in\cite{hoffman2018cycada} and \cite{sun_ECCV2016} have used Cycle consistency~\cite{zhu_CVPR2017} loss and deep coral loss~\cite{sun_DACVA2017,sun_2016AAAI} for ensuring the closeness of source and target domain respectively.
Deep Bayesian models have been used to play an important role in the estimation of the deep model uncertainty. The Bayesian formulation in domain adaptation natural language processing has been proposed in~\cite{finkel2009hierarchical}. In~\cite{gal_icml2016dropout}, it has been justified that dropout can also work as an approximation of the deep Bayesian network. Works on the uncertainty estimation have been reported in~\cite{Gal2016Uncertainty,kendall2017uncertainties}.  In~\cite{malinin2018predictive} predictive uncertainty has been calculated over the prior networks. Uncertainty in the ensemble model along with adversarial training has been discussed in~\cite{lakshminarayanan_nips2017}. Another work on the Bayesian uncertainty estimation has been reported in the~\cite{teye2018bayesian}.

Attention-based networks have been widely applied in many computer vision applications such as image captioning~\cite{xu2015show,you2016image}, visual question answer~\cite{yang2016stacked,Patro_2018_CVPR,kazemi2017show} and speech recognition~\cite{chorowski2015attention}. The advantage of the attention model is that it helps to learn some set of weights over a set of representation input which has relatively more importance than others. Recently~\cite{vaswani2017attention} showed that the attention mechanism can also be achieved by dispensing with recurrence and convolutions. A recent work~\cite{Kang_2018_ECCV} addresses the domain adaptation problem by obtaining the synthetic source and target images from CycleGAN~\cite{zhu_CVPR2017}, and then aligned the attention map of all the pairs. 
\vspace{-0.2cm}

\section{Methodology}
In the unsupervised domain adaptation problem, the source dataset~$\mathcal{D}_s= (x_i^s,y_i^s)$ consists of data sample~($x_i^s$) with corresponding label~($y_i^s$) where $\mathcal{D}_s \in P_s $ and the target dataset ~$\mathcal{D}_t(x_i^t)$ consists of unlabeled data samples~($x_i^t$) where $\mathcal{D}_t \backsim P_t $.  $P_s$ and $P_t$ are the source and target distributions. We further assume that both the domains are complex and unknown.
For solving this task, we are following the adversarial domain adaptation framework, where a discriminator is trained to learn domain invariant features domain invariant while a classifier is trained to learn class discriminative features.  
 In this paper, we are proposing a discriminator certainty based domain adaption model represented in the Figure~\ref{fig:attention_model}, which consists of three major modules: Feature extractor, Bayesian Classifier, and Bayesian Discriminator. The feature extractor is pretrained on the Imagenet dataset, while both the classifier and discriminator are Bayesian neural networks (BNN). We have followed the approach defined in ~\cite{gal_icml2016dropout,kendall2017uncertainties,kendall2017multi} for transforming deep neural networks into BNNs.
\subsection{Bayesian Classifier}
Bayesian framework is one of the efficient ways to predict uncertainty. Gal et.al~\cite{Gal2016Uncertainty} has shown that by applying dropout after every fully connected (fc) layer, we can perform probabilistic inference for deep neural networks. Hence we have followed a similar approach for defining the classifier.
For estimating uncertainty, similar to~\cite{kendall2017uncertainties,WinNT}, we trained the classifier to output class probabilities along with aleatoric uncertainty (data uncertainty). The predictive uncertainty includes both model uncertainty and data uncertainty, where model uncertainty results from uncertainty in model parameters. Estimation of aleatoric uncertainty for the classifier makes the features more robust for prediction, and estimation of predictive uncertainty provides a tool for visualizing model's predictions. 

For the input sample $x_i$, the feature extractor $G_f$ outputs features $\mathbf{f_i}$, represented by $\mathbf{f_i} = G_f(\mathbf{x_i})$. The predicted class logits $y_i^c$ and aleatoric uncertainty $v_i^c$ are obtained as:
\begin{align}
    && y_i^c &= G_{cy}(G_c(\mathbf{f_i})) , &v_i^c = G_{cv}(G_c(\mathbf{f_i}))
\end{align}
 
where $G_{cy}$   and $G_{cv}$ are the logits and aleatoric uncertainty prediction modules of the classifier $G_c$ respectively.
The classification loss for predicted logits is defined as: 
\begin{equation}
    \mathcal{L}_{cy}=\frac{1}{n_s} \sum_{x_i \in \mathcal{D}_s} \mathcal{L}(y_i^c,y_i)
\end{equation}
where $\mathcal{L}$ is the cross entropy loss function and  $y_i$ is the true class label for the input $x_i$. The total number of data samples in the source domain is denoted as $n_s$. The classifier aleatoric loss $ \mathcal{L}_{cv}$ for predicted uncertainty $v_i^c$ is defined as:
\begin{equation}
\begin{split}
    \hat{y}_{i,t}^c = y_i^c+\sigma_{i}^c* \epsilon_t ,\ \;\;\; \epsilon_t \sim \mathcal{N}(0, I) \\
    \mathcal{L}_{cv}= -\frac{1}{n_s} \sum_{x_i \in \mathcal{D}_s}\log  \frac{1}{T} \sum_t \mathcal{L} (\hat{y}_{i,t}^c,y_i) 
\end{split}
\end{equation}
 where $\sigma_{i}^c$ is the standard deviation, $v_{i}^c = (\sigma_{i}^c)^2$. The classifier is trained by jointly minimizing both the classification loss $\mathcal{L}_{cy}$ and  aleatoric loss $\mathcal{L}_{cv}$.

\subsection{Bayesian Discriminator}
In the proposed method, the discriminator is also modeled in the Bayesian framework similar to the Bayesian classifier. The uncertainty in the discriminator network implies the region where it is uncertain about its prediction about the domain. The uncertainty estimation of the discriminator can guide the feature extractor more efficiently for domain adaptation. All real-world images contain some type of aleatoric uncertainty or noise. These regions which contain aleatoric uncertainty, are not adaptable. 
By aligning these uncertain regions, we are corrupting the feature representation, thus confusing the classifier during predictions for the target domain. So, by estimating aleatoric uncertainty, the discriminator is avoiding the learning of feature representations for these regions, which also reduces negative transfer~\cite{pei_arxiv2018}. The negative transfer introduces false alignment of the mode of two distributions across domains, which needs to be prevented during adaptation.
Similarly, the predictive uncertainty tells us about the model's incapability to classify the domains, as the discriminator is not sure about the domain. Predictive uncertainty occurs in the region where either it is already adapted, or there is noise which corresponds to aleatoric uncertainty.
We obtain the discriminator predicated logits and variance using the following equations
\begin{align}
    && y_i^d &= G_{dy}(G_d(\mathbf{f_i})) , & v_i^d = G_{dv}(G_d(\mathbf{f_i}))
\end{align}
where $G_{dy}$ and $G_{dv}$ predict domain class logits $y_i^d$ and domain aleatoric uncertainty $v_i^d$ respectively using features from $G_d$.
The domain classification loss $ \mathcal{L}_{dy}$ is defined as:
\begin{equation}
    \mathcal{L}_{dy}=\frac{1}{n_s+n_t} \sum_{x_i \in \mathcal{D}_s \cup \mathcal{D}_t} \mathcal{L}(y_i^d,d_i)
\end{equation}

where $ \mathcal{L}$ is the cross entropy loss function, $d_i$ is the true domain of the image, and  $n_s$ and $n_t$ are the number of source and target samples. 
The domain label  $d_i$ is defined to be $0$ if $x_i \in \mathcal{D}_s$ and $1$ if $x_i \in \mathcal{D}_t$.
The discriminator aleatoric loss $ \mathcal{L}_{dv}$ is defined as:
\begin{equation}
\begin{split}
    \hat{y}_{i,t}^d = y_i^d+\sigma_{i}^d* \epsilon_t ,\ \;\;\; \epsilon_t \sim \mathcal{N}(0, I) \\
    \mathcal{L}_{dv}= -\frac{1}{n_s + n_t} \sum_{x_i \in  \mathcal{D}_s \cup \mathcal{D}_t} \log  \frac{1}{T} \sum_t \mathcal{L} (\hat{y}_{i,t}^d,d_i) 
\end{split}
\end{equation}
 where $v_{i}^d = (\sigma_{i}^d)^2$. Discriminator is trained by jointly minimizing both the domain classification loss $\mathcal{L}_{dy}$ and discriminator aleatoric loss $\mathcal{L}_{dv}$.

\subsection{Certainty Based Attention}
\begin{figure*}[!ht]
     \centering
       \includegraphics[scale=0.24]{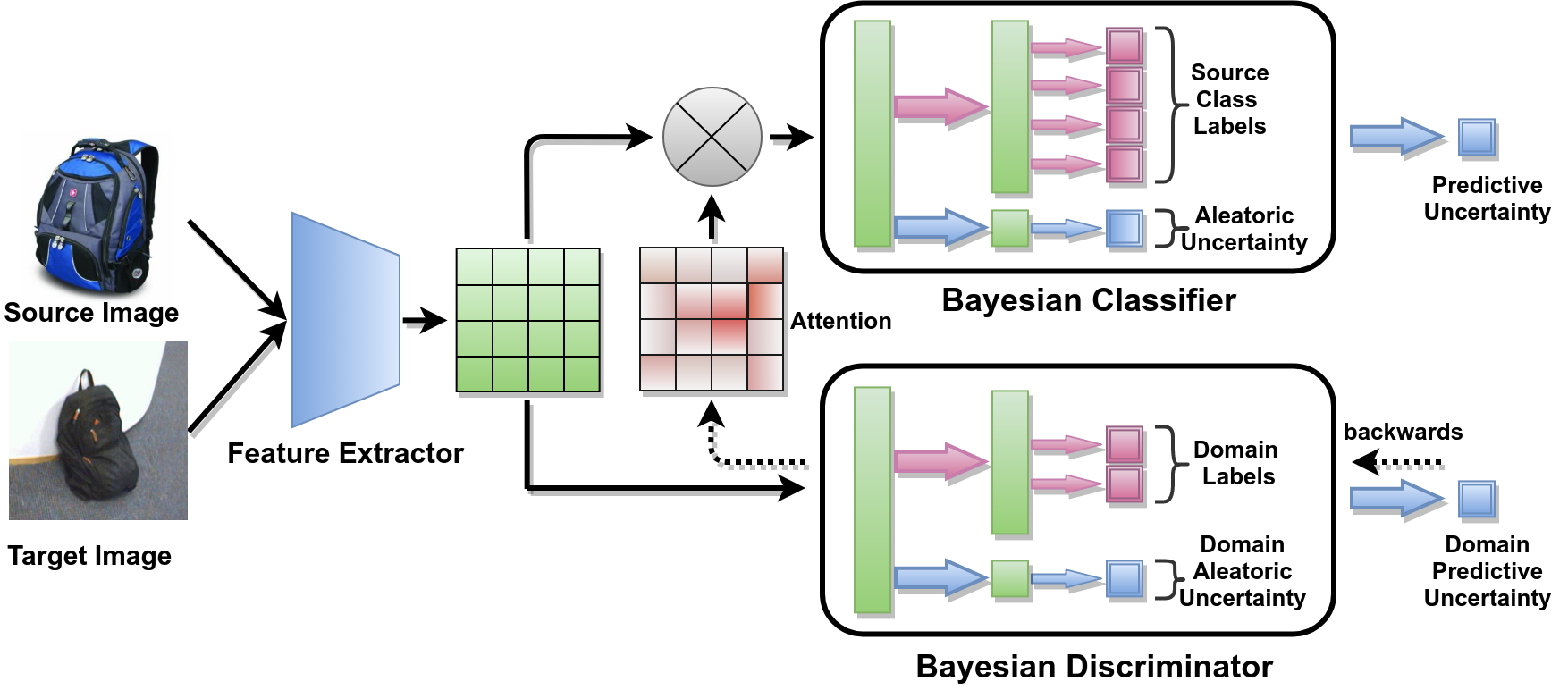}
  \caption{The architecture of Certainty based Attention for Domain Adaptation (CADA), consists of a shared feature extractor, Bayesian classifier and Bayesian discriminator where both the classifier and discriminator predict the variance value along with the prediction score. Discriminator's predictive or aleatoric uncertainty is used to highlight the regions where the discriminator is certain about its predictions.}
      \label{fig:attention_model}
      \vspace{-0.7em}
 \end{figure*}
Uncertainty estimation of the discriminator can help in identifying those regions which can be adapted, cannot be adapted, or already adapted.
The regions which are already aligned will confuse the discriminator for predicting the domain. Hence discriminator will be highly uncertain on these regions. The discriminator will also be highly uncertain on the regions containing aleatoric uncertainty, and these regions can't be adapted. Uncertainty estimation can also help to identify regions where discriminator is certain or the regions which can be further aligned.

 In most of the datasets, the discriminator can easily discriminate between the source and target images by only attending on the background during the initial phase of the training. Hence, the discriminator will be more certain in these regions, which results in easier adaptation of background regions after some adversarial training. But the foreground regions are difficult to adapt, as foreground varies a lot across all the classes and images. Therefore for most of the span during training, the discriminator will be certain on the transferable regions of the foreground. Thus, if the classifier attends to certain regions of the discriminator, it will focus more on the transferable regions of the foreground during training. 
 
The Certainty Attention based Domain Adaption (CADA) model is shown in Figure~\ref{fig:attention_model}. In the proposed work, we propose the two variants of CADA: aleatoric certainty based attention (CADA-A) and predictive certainty based attention (CADA-P). 
 

\vspace{-0.4cm}
\subsubsection{Aleatoric Certainty based Attention}
The aleatoric uncertainty ($v_i^d$) of the domain discriminator occurs due to corruption or noise in the regions. These regions are not suited for the adaptation as well as for object classification and should be less focused as compared to the certain regions for the classification task. For identifying the aleatoric uncertain regions, we compute the gradient of aleatoric uncertainty with respect to the features $f_i$.
These gradients~($\frac{\partial{v_{i}^d}}{\partial{f_i}}$) will flow back through the gradient reversal layer, and will correspond to the gradients of aleatoric certainty, i.e.,~$-\frac{\partial{v_{i}^d}}{\partial{f_i}}$.
\vspace{-0.4cm}
\begin{equation}
    p_i = f_i * -\frac{\partial{v_{i}^d}}{\partial{f_i}}
\end{equation}
\vspace{-0.1cm}
Therefore the positive sign of the product of features and gradients of the aleatoric certainty denotes the positive influence of aleatoric certainty on these features. i.e. the discriminator is certain on these regions.
\vspace{-0.2em}
\begin{equation}
    a_i = \text{ReLU}(p_i) + c*\text{ReLU}(-p_i) \label{eqn:wi}
\end{equation}
\vspace{-0.2em}
For obtaining the regions where the discriminator is certain, the product $p_i$ is passed through a ReLU function. But for ignoring the negative values that correspond to uncertain regions, $-p_i$ is again passed through a ReLU function. This is then multiplied with a large number $c$, such that after applying softmax over $a_i$ all negative values of $p_i$ become zero and all the positive values are normalized.  
\begin{equation}
    w_i= (1-v_{i}^d) *\text{Softmax}(a_i)  \label{eqn:w_d}
\end{equation}

\vspace{-0.6em}
To focus on more attentive (certain) regions, we follow the residual setting~\cite{long_NIPS2016} for obtaining the effective weighted features $h_i$. Images with high aleatoric uncertainty (lower certainty) should be less attentive, and it is obtained by multiplying the normalized softmax attention value to its certainty value $(1-v_{i}^d)$ using the Eq.~\ref{eqn:w_d}. The weighted feature $h_i$ is generated as follows:
\begin{equation}
    h_i=f_i*(1+w_i)
\end{equation}
\subsubsection{Predictive Certainty based Attention}
The predictive uncertainty measures the model's capability for prediction. It occurs in the regions which are either already domain invariant or which contain noise. The regions corresponding to discriminator's predictive certainty are domain transferable regions and should be attended during classification. We follow the approach proposed in~\cite{kendall2017uncertainties} for computing the predictive uncertainty of discriminator. It is obtained by the entropy of the average class probabilities of the Monte Carlo (MC) samples.
We sample weights from $\mathbf{G_d}$, and perform MC simulations over domain probabilities $p(y_{i, t}^d)$ and aleatoric uncertainty $v_{i, t}^d$ for estimating predictive uncertainty $\text{Var }^d(f_i)$.
\begin{equation}
    g_{i,t} = G_d^t(\mathbf{f_i}) ,\ \;\;\;  G_d^t \sim \mathbf{G_d}
\end{equation}
Using the sampled model, we calculate domain probabilities and aleatoric uncertainty.
\begin{align}
    &&     p(y_{i, t}^d) &= \text{Softmax}(G_{dy}(g_{i,t})) , & v_{i, t}^d = G_{dv}(g_{i,t})
\end{align}
\begin{equation}
    H(y_{i, t}^d) = - \sum_{c=1}^2 p(y_{i, t}^d=c) *\log (p(y_{i, t}^d=c))
\end{equation}
\begin{equation}
\text{Var }^d(f_i) \approx \frac{1}{T} \sum_{t=1}^T \Big( v_{i, t}^d + H(y_{i, t}^d) \Big)
\end{equation}
where $H(y_{i, t}^d)$ is the Entropy of the $p(y_{i, t}^d)$. For identifying the predictive uncertain regions, we compute gradients of the predictive uncertainty $\text{Var }^d(f_i)$ of the discriminator with respect to the features $f_i$ i.e. $\frac{\partial{\text{Var }^d(f_i)}}{\partial{f_i}}$, and negative of these gradients which returns through the gradient reversal layer will correspond to the gradients of the predictive certainty, i.e., $-\frac{\partial{\text{Var }^d(f_i)}}{\partial{f_i}}$.
\begin{equation}
    p_i = f_i*-\frac{\partial{\text{Var }^d(f_i)}}{\partial{f_i}}
\end{equation}
\begin{equation}
   a_i = \text{ReLU}(p_i) + c*\text{ReLU}(-p_i) \label{eqn:wj}
\end{equation}
\begin{equation}
    w_i= (1-\text{Var }_i^d(f_i)) *\text{Softmax}(a_i)
\end{equation}
Similar to aleatoric certainty based attention, for obtaining the predictive certain regions we apply ReLU function to $p_i$ and ignoring its negative values (corresponds to uncertain regions), a   ReLU function is applied to negative of $p_i$, and multiply it with a large number $c$ using Eq.~\ref{eqn:wj}. After applying the Softmax, the features that are activated by the predictive uncertainty will have zero weight and for features that are highly activated by the predictive certainty will get more weight.
The residual weighted features are obtained by following equations 
\vspace{0.5em}
\begin{equation}
    h_i=f_i*(1+w_i)
\end{equation}
Therefore the images with high predictive uncertainty have a lower value of $w_i$, have less attention, while images have high predictive certainty have a high value of $w_i$, produce high attentive features. This will ensure that already adapted regions or non- adaptive region (both cases have high uncertainty) have a lower attention value.

\subsection{Training Algorithm}
We employ Certainty Attention based Domain Adaption (CADA) models for solving the task of unsupervised domain adaptation. Both the CADA-P and CADA-A models jointly learn domain invariant and class invariant features, by focusing the classifier's attention on the discriminator's certain regions. So, the class probabilities $y_i^c$ and aleatoric uncertainty $v_i^c$ for the classifier will be estimated using weighted feature $h_i$.
\begin{table*}
\centering
\caption {Classification accuracy (\%) on \textit{Office-31} dataset for unsupervised domain adaptation (AlexNet\cite{krizhevsky_NIPS2012})} 
\begin{tabular}{cccccccc}
\toprule
\textbf{Method }& A $\rightarrow$ W & D$\rightarrow$ W &  W $\rightarrow$ D &A $\rightarrow$ D & D $\rightarrow$ A & W $\rightarrow$ A & Average \\ 
  \midrule
 Alexnet\cite{krizhevsky_NIPS2012}  & 60.6 $\pm$ 0.4 & 95.0$\pm$ 0.2  & 99.5$\pm$ 0.1 &64.2 $\pm$ 0.3& 45.5$\pm$ 0.5  & 48.3$\pm$ 0.5 & 68.8\\
  MMD\cite{tzeng_arxiv2014} & 61.0 $\pm$ 0.5 & 95.0$\pm$ 0.3  & 98.5$\pm$ 0.3 &64.9 $\pm$ 0.4& 47.2$\pm$ 0.5  & 49.4$\pm$ 0.4&69.3 \\ 
  RTN\cite{long_NIPS2016} & 73.3 $\pm$ 0.3 & 96.8 $\pm$ 0.2 & 99.6 $\pm$ 0.1& 71.0$\pm$ 0.2 & 50.5$\pm$ 0.3  & 51.0$\pm$ 0.1 & 74.1\\ 
  DAN\cite{long_ICML2015} & 68.5 $\pm$ 0.4 & 96.0 $\pm$ 0.3 & 99.0 $\pm$ 0.2 & 66.8$\pm$ 0.2 & 50.0$\pm$ 0.4  & 49.8$\pm$ 0.3 & 71.7 \\ 
  GRL \cite{ganin_ICML2015} & 73.0 $\pm$ 0.5 & 96.4 $\pm$ 0.3  & 99.2 $\pm$ 0.3 & 72.3 $\pm$ 0.3 & 52.4$\pm$ 0.4  & 50.4$\pm$ 0.5 & 74.1\\
  JAN \cite{long_ICML2017} & 75.2 $\pm$ 0.4 & 96.6 $\pm$ 0.2  & 99.6 $\pm$ 0.1 & 72.8 $\pm$ 0.3 & 57.5$\pm$ 0.2  & 56.3$\pm$ 0.2 & 76.3\\
  CDAN\cite{long_arxive2017conditional} & 77.9 $\pm$ 0.3 & 96.9 $\pm$ 0.2 &  100.0 $\pm$ 0 & 74.6 $\pm$ 0.2 & 55.1$\pm$ 0.3  &  57.5$\pm$ 0.4 & 77.0\\ 
 MADA\cite{pei_arxiv2018} & 78.5 $\pm$ 0.2 & 99.8 $\pm$ 0.1 &  100.0 $\pm$ 0 & 74.1 $\pm$ 0.1 & 56.0$\pm$ 0.2  & 54.5$\pm$ 0.3 & 77.1\\ 
 \midrule
  CADA-W & 82.3$\pm$ 0.3 & 99.2$\pm$ 0.1 & 99.6 $\pm$ 0.1& 75.9$\pm$ 0.2& 57.7$\pm$ 0.1 & 53.3$\pm$ 0.2 & 78.0\\ 
CADA-A & \textbf{84.1 $\pm$ 0.2} & 99.2 $\pm$ 0.2 &  99.8$\pm$ 0.2 & 77.3 $\pm$ 0.1 & \textbf{61.3$\pm$ 0.2}  & 54.1$\pm$ 0.3 & 79.3\\ 
CADA-P & 83.4 $\pm$ 0.2 & \textbf{99.8 $\pm$ 0.1} &  \textbf{100.0 $\pm$ 0} & \textbf{80.1 $\pm$ 0.1} & 59.8$\pm$ 0.2  & \textbf{59.5$\pm$ 0.3} & \textbf{80.4}\\
 \bottomrule 
\end{tabular}
  \label{tbl:alex_office_table}
 \end{table*}
\begin{align}
    && y_i^c &= G_{cy}(G_c(\mathbf{h_i})) , & v_i^c = G_{cv}(G_c(\mathbf{h_i}))
\end{align}
The final objective function $J$ for optimizing both the models is defined as:
\begin{equation}
\begin{split}
J = \mathcal{L}_{cy} +\mathcal{L}_{cv} - \lambda*(\mathcal{L}_{dy}+\mathcal{L}_{dv})
\end{split}
\end{equation}
where $\lambda$ is a trade-off parameter between classifier and discriminator. The optimization problem is to find the parameters $\hat{\theta}_f, \hat{\theta}_c, \hat{\theta}_{cy}, \hat{\theta}_{cv}, \hat{\theta}_d, \hat{\theta}_{dy}, \hat{\theta}_{dv}$ that jointly satisfy: 
\begin{equation*}
    (\hat{\theta}_f, \hat{\theta}_c, \hat{\theta}_{cy}, \hat{\theta}_{cv}) =\underset{\begin{subarray}{c}
  \theta_f, \theta_c, \\
  \theta_{cy}, \theta_{cv}
  \end{subarray}}{\text{arg min }} J(\theta_f, \theta_c, \theta_{cy}, \theta_{cv}, \theta_{d}, \theta_{dy}, \theta_{dv})
\end{equation*}
\vspace{-0.3cm}
\begin{equation*}
    (\hat{\theta}_d, \hat{\theta}_{dy}, \hat{\theta}_{dv}) =\underset{\theta_d, \theta_{dy}, \theta_{dv}}{\text{arg max}}J(\theta_f, \theta_c, \theta_{cy}, \theta_{cv}, \theta_{d}, \theta_{dy}, \theta_{dv})
\end{equation*}
The implementation details are provided in supplementary material, and other details are provided in the project page~\footnote{https://delta-lab-iitk.github.io/CADA/}
\section{Experiments and Results}
\begin{table*}[!]
\centering
\caption {Classification accuracy (\%) on \textit{Office-31} dataset for unsupervised domain adaptation (ResNet-50~\cite{he2016deep})} 
\begin{tabular}{cccccccc}
 \toprule
  \textbf{Method }& A $\rightarrow$ W & D$\rightarrow$ W &  W $\rightarrow$ D &A $\rightarrow$ D & D $\rightarrow$ A & W $\rightarrow$ A & Average \\ 
  \midrule
 ResNet-50\cite{he2016deep} &   68.4$\pm$0.2   &    96.7$\pm$0.1  &    99.3$\pm$0.1    &    68.9$\pm$0.2   &    62.5$\pm$0.3  &    60.7$\pm$ 0.3 &   76.1\\
 DAN\cite{long_ICML2015} &   80.5$\pm$0.4   &    97.1$\pm$0.2   &    99.6$\pm$0.1    &    78.6$\pm$0.2    &    63.6$\pm$0.3   &    62.8$\pm$0.2  &   80.4\\
 RTN\cite{long_NIPS2016} &   84.5$\pm$0.2   &    96.8$\pm$0.1  &    99.4$\pm$0.1    &    77.5$\pm$0.3    &    66.2$\pm$0.2   &    64.8$\pm$0.3  &   81.6\\
 DANN\cite{ganin_ICML2015} &   82.0$\pm$0.4   &    96.9$\pm$0.2   &    99.1$\pm$0.1    &    79.7$\pm$0.4    &    68.2$\pm$0.4   &    67.4$\pm$0.5  &  82.2 \\
 ADDA \cite{tzeng_CVPR2017}&   86.2$\pm$0.5   &    96.2$\pm$0.3   &    98.4$\pm$0.3    &    77.8$\pm$0.3    &    69.5$\pm$0.4   &    68.9$\pm$0.5  &  82.9 \\
 JAN\cite{long_ICML2017} &   85.4$\pm$0.3   &    97.4$\pm$ 0.2  &    99.8$\pm$0.2    &    84.7$\pm$ 0.3   &    68.6$\pm$0.3   &    70.0$\pm$0.4  &  84.3 \\
 MADA\cite{pei_arxiv2018} &  90.0$\pm$ 0.1  &   97.4 $\pm$0.1   &    99.6$\pm$ 0.1   &   87.8$\pm$0.2    &    70.3$\pm$0.3   &    66.4$\pm$0.3  &  85.2 \\
 SimNet\cite{pinheiro_cvpr2017unsupervised} & 88.6 $\pm$0.5  &  98.2$\pm$0.2   &    99.7$\pm$0.2    &    85.2$\pm$0.3    &   73.4$\pm$0.8  &   71.6$\pm$0.6 &  86.2 \\
 GTA\cite{sankaranarayanan_cvpr2018learning} &   89.5$\pm$0.5   &    97.9$\pm$0.3   &    99.8$\pm$0.4    &   87.7$\pm$0.5    &    72.8$\pm$0.3   &    71.4$\pm$0.4  &  86.5 \\
 DAAA~\cite{ Kang_2018_ECCV} &  86.8$\pm0.2$   &    99.3$\pm$0.1   &    100.0$\pm$0.0    &    88.8$\pm$0.4    &    \textbf{74.3$\pm$0.2}  &   \textbf{73.9 $\pm$0.2}  & 87.2  \\
  CDAN\cite{long_arxive2017conditional} &   93.1$\pm0.1$   &    98.6$\pm$0.1   &    100.0$\pm$0.0    &    93.4$\pm$0.2    &    71.0$\pm$0.3  &   70.3 $\pm$0.3  & 87.7  \\
 \midrule
  CADA-W & 93.9$\pm$ 0.1 & 99.1 $\pm$ 0.2 &  99.6$\pm$ 0.2 & 93.2$\pm$ 0.3 & 68.9$\pm$ 0.1  & 68.3$\pm$ 0.2 & 87.2\\ 
 CADA-A & 96.8 $\pm$ 0.2 & 99.0$\pm$0.1 &  99.8 $\pm$0.1 & 93.4$\pm$0.1 & 71.7$\pm$0.2  & 70.5$\pm$0.3 & 88.5\\ 
CADA-P & \textbf{97.0 $\pm$ 0.2} & \textbf{99.3 $\pm$0.1} &  \textbf{100.0$\pm$0} & \textbf{95.6$\pm$0.1} & 71.5$\pm$0.2  & 73.1$\pm$0.3 & \textbf{89.5}\\
\bottomrule
\end{tabular}
  \label{tbl:res_office_table}
 \end{table*}
 
 
 \begin{table*}[!]
 \centering
\caption {Classification accuracy (\%) on \textit{Office-Home} dataset for unsupervised domain adaptation (ResNet-50~\cite{he2016deep})} 

\setlength\tabcolsep{1.15pt}%
\begin{tabular}{*{200}{c}}
 \toprule
  \textbf{Method}& Ar$\rightarrow$Cl & Ar$\rightarrow$Pr &  Ar$\rightarrow$Rw & Cl$\rightarrow$Ar & Cl$\rightarrow$Pr & Cl$\rightarrow$Rw & Pr$\rightarrow$Ar & Pr$\rightarrow$Cl & Pr$\rightarrow$Rw & Rw$\rightarrow$Ar & Rw$\rightarrow$Cl& Rw$\rightarrow$Pr & Avg \\ 
  \midrule
 ResNet-50\cite{he2016deep} & 34.9 & 50.0& 58.0 & 37.4 & 41.9 & 46.2 & 38.5 &31.2 & 60.4 & 53.9 & 41.2 & 59.9 & 46.1\\
 DAN\cite{long_ICML2015} & 43.6 & 57.0 & 67.9 & 45.8 & 56.5 & 60.4 & 44.0 & 43.6 & 67.7 & 63.1 & 51.5 & 74.3 & 56.3\\
DANN\cite{ganin_ICML2015} &45.6&59.3 & 70.1&47.0 &58.5 &60.9 &46.1 &43.7 &68.5 &63.2 &51.8 &76.8 &57.6 \\
JAN\cite{long_ICML2017} & 45.9 & 61.2 &68.9 &50.4 & 59.7 & 61.0 &45.8 &43.4&70.3 &63.9 &52.4 &76.8 &58.3 \\
CDAN\cite{long_arxive2017conditional} & 50.6 & 65.9 & 73.4 & 55.7 & 62.7 & 64.2 & 51.8 & 49.1 & 74.5 & 68.2 & 56.9 & 80.7 & 62.8\\
\midrule
CADA-A & \textbf{56.9} & 75.4 & 80.2 & \textbf{61.7} & 74.6 & 74.9 & 62.9 &  54.4 & \textbf{80.9} & \textbf{74.3} & 61.1 & \textbf{84.4 }& 70.1\\
CADA-P & \textbf{56.9} & \textbf{76.4} & \textbf{80.7} & 61.3 & \textbf{75.2} & \textbf{75.2} & \textbf{63.2} &  \textbf{54.5} & 80.7 & 73.9 & \textbf{61.5} & 84.1 & \textbf{70.2}\\
 \bottomrule
\end{tabular}
  \label{tbl:home_office}
 \end{table*}
 
\begin{table}[!]
\centering
\caption {Classification accuracy (\%) on \textit{ImageCLEF} dataset for unsupervised domain adaptation (ResNet-50~\cite{he2016deep})} 
\setlength\tabcolsep{1pt}%
\begin{tabular}{cccccccc}
 \toprule
  \textbf{Method }& I $\rightarrow$ P & P$\rightarrow$ I &  I $\rightarrow$ C &C $\rightarrow$ I & C $\rightarrow$ P & P $\rightarrow$ C & Avg \\ 
  \midrule
ResNet~\cite{he2016deep} & 74.8 & 83.9 & 91.5 & 78.0 & 65.5 & 91.2 & 80.7 \\
DAN\cite{long_ICML2015} & 75.0 & 86.2 & 93.3 & 84.1 & 69.8 & 91.3 & 83.3 \\
RTN\cite{long_NIPS2016} & 75.6 & 86.8 & 95.3 & 86.9 & 72.7 & 92.2 & 84.9 \\
GRL~\cite{ganin_ICML2015} & 75.0 & 86.0 & 96.2 & 87.0 & 74.3 & 91.5 & 85.0 \\
JAN\cite{long_ICML2017} & 76.8 & 88.0 &94.7 & 89.5 & 74.2 & 91.7 & 85.8 \\ 
MADA~\cite{pei_arxiv2018} & 75.0 & 87.9 & 96.0 & 88.8 & 75.2 & 92.2 & 85.8 \\
CDAN\cite{long_arxive2017conditional} & 77.2 & 88.3 & \textbf{98.3} & 90.7 & 76.7 & 94.0 & 87.5  \\
 \midrule
 CADA-A & \textbf{78.0} & \textbf{91.5} & 96.3 & 91.0 & 77.1 & 95.3 & 88.2\\ 
CADA-P & \textbf{78.0} & {90.5} & 96.7 & \textbf{92.0} &  \textbf{77.2}  &  \textbf{95.5} & \textbf{88.3} \\
\bottomrule
\end{tabular}
  \label{tbl:imageclef}
 \end{table}
 
\begin{figure*}[!]
\begin{subfigure}{0.19\textwidth}
  \centering
  \includegraphics[scale=0.15]{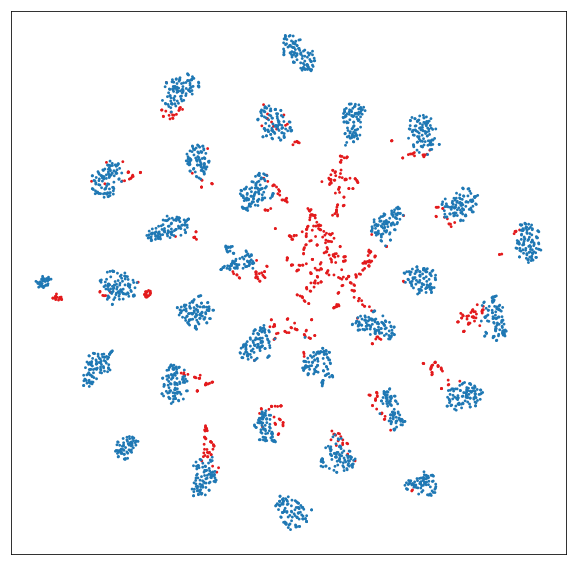}
  \caption{ResNet}
  \label{fig:sfig1111}
\end{subfigure}
\begin{subfigure}{0.19\textwidth}
  \centering
  \includegraphics[scale=0.15]{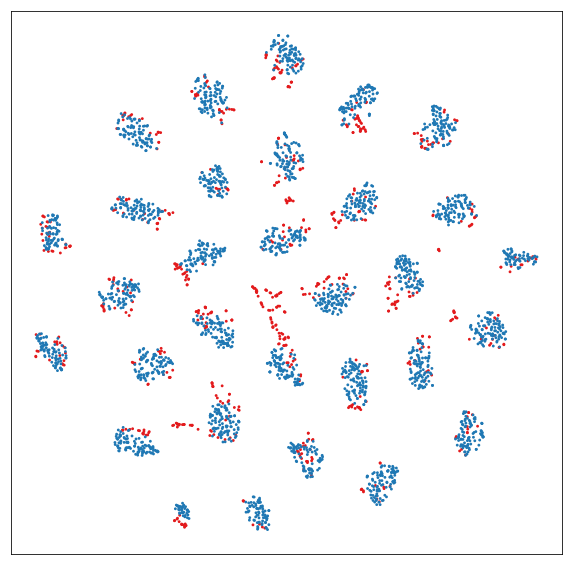}
  \caption{DANN}
  \label{fig:sfig22}
\end{subfigure}
\begin{subfigure}{0.19\textwidth}
  \centering
  \includegraphics[scale=0.15]{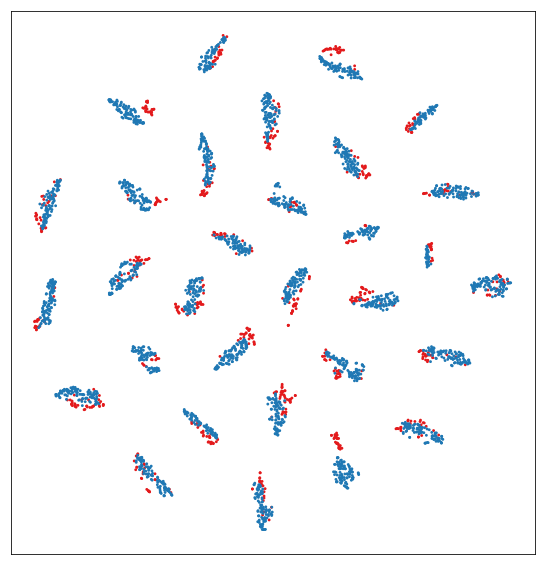}
  \caption{CADA-A}
  \label{fig:sfig3}
\end{subfigure}
\begin{subfigure}{0.19\textwidth}
  \centering
  \includegraphics[scale=0.15]{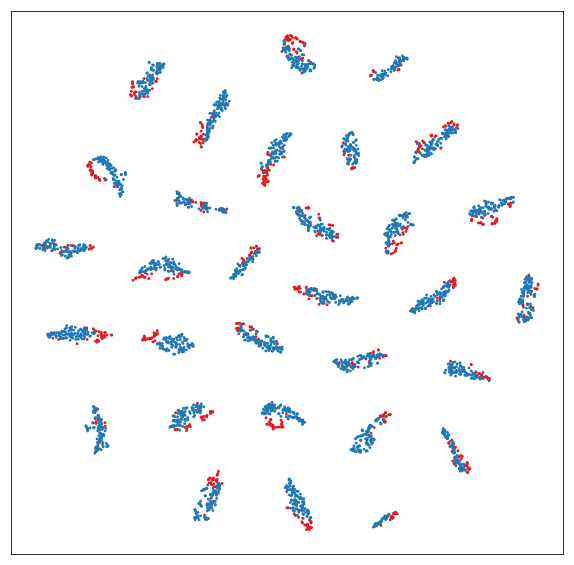}
  \caption{CADA-P}
  \label{fig:sfig44}
\end{subfigure}
\begin{subfigure}{0.14\textwidth}
  \centering
  \includegraphics[scale=0.165]{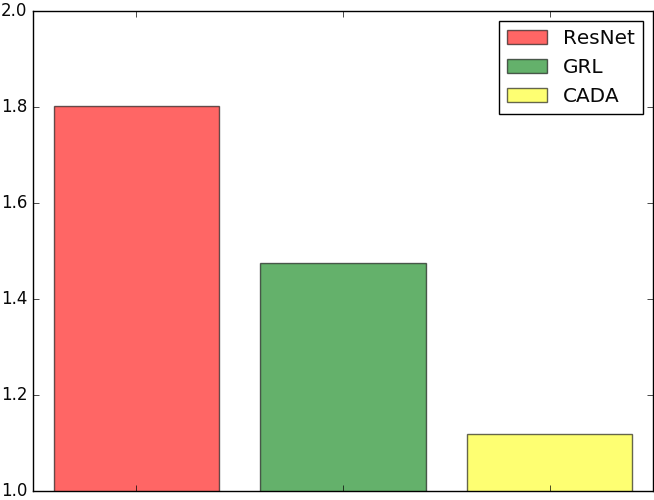}
  \caption{Proxy distance}
  \label{fig:sfig5}
\end{subfigure}
\caption{The t-SNE visualization of representations learned by (a) ResNet, (b) DANN, (c) CADA-A, and (d) CADA-P ({\color{blue} blue}: A; {\color{red} red}: W), 
(e) shows Proxy$\mathcal{A}$-distance for A $\rightarrow$ W task for method Reset~\cite{he2016deep}, GRL~\cite{ganin_ICML2015} and the proposed model CADA-P}
\label{fig:tsne}
\vspace{-0.5em}
\end{figure*}
\subsection{Datasets}

\begin{figure}
\vspace{-0.5em}
\begin{subfigure}{.24\textwidth}
  \centering
  \includegraphics[width=1\linewidth]{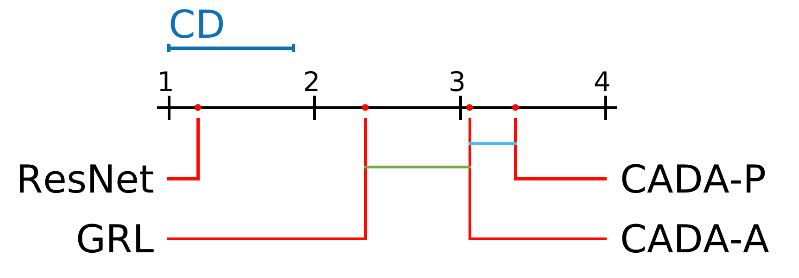}
  \caption{ A $\rightarrow$ W }
  \label{fig:sfig10}
\end{subfigure}%
\begin{subfigure}{.24\textwidth}
  \centering
  \includegraphics[width=1\linewidth]{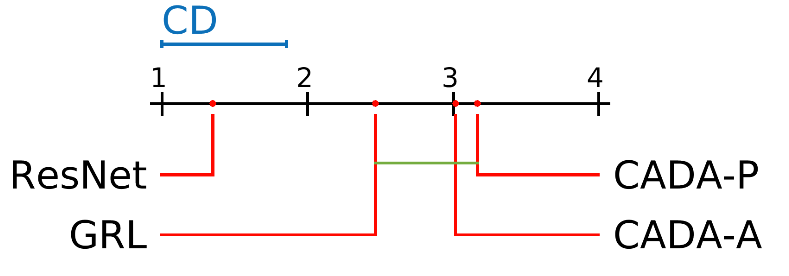}
  \caption{A $\rightarrow$ D}
  \label{fig:sfig23}
\end{subfigure}
 \caption{Analysis of statistically significant difference for A$\rightarrow$W  and A $\rightarrow$D in ResNet~\cite{he2016deep}, GRL~\cite{ganin_ICML2015}, CADA-A, and CADA-P methods, with a significance level of 0.05.}
 \label{fig:ssa1}
 \vspace{-1.5em}
\end{figure}
\textbf{Office-31 Dataset}
The Office-31~\cite{saenko_ECCV2010} consists of three domains: Amazon, Webcam and DSLR with 31 classes in each domain.
There are 2817 images in Amazon (A) domain, 795 images in Webcam (W) and 498 images are in DSLR (D) domain makes total 4,110 images.
To enable unbiased evaluation, we evaluate all methods on 6 transfer tasks A$\rightarrow$W, D$\rightarrow$A, W$\rightarrow$A, A $\rightarrow$D, D$\rightarrow$ W and W $\rightarrow$ D.
\vspace{0.8em}

\textbf{Office-Home Dataset}
We also evaluated our model on the Office-Home dataset~\cite{venkateswara_cvpr2017deep} for unsupervised domain adaptation. This dataset consists of four domains: Art (Ar), Clip-art (Cl), Product (Pr) and Real-World (Rw). Each domain has common 65 categories and total 15,500 images. 
We evaluated our model by considering all the 12 adaptation tasks.
The performance is reported in the Table~\ref{tbl:home_office}.
\vspace{0.8em}

\textbf{ImageCLEF Dataset}
ImageCLEF-2014 dataset consists of three datasets: Caltech-256 (C), ILSVRC 2012 (I), and Pascal VOC 2012~(P). There are 12 common classes and total 600 images in each domain. We evaluate our model on all the 6 transfer tasks and results are reported in Table~\ref{tbl:imageclef}.
\subsection{Results}
\vspace{-0.5em}
Following the common setting in unsupervised domain adaption, we used the pre-trained Alexnet~\cite{krizhevsky_NIPS2012} and pre-trained ResNet~\cite{he2016deep} architecture as our base model.
For Office-31 dataset, results are reported in the Table~\ref{tbl:alex_office_table} and Table~\ref{tbl:res_office_table}. From the table, it is clear that the proposed CADA  outperforms the other methods on most transfer tasks, where CADA-P is the top-performing variant for both Alexnet and Resnet model. On average, we obtain improvements of 3.3\% and 1.8\% over the state of the art methods as it can be seen that the difference between other methods are usually less than 1\% and therefore this amount of improvement is fairly significant. In some cases such as, in Amazon-Webcam (A-W) we obtain almost 4\% improvement over the state of the art method. Note that for DSLR-Amazon (D-A) and Webcam-Amazon (W-A), we do not obtain state of the art. A very recent work  \cite{Kang_2018_ECCV} obtain state of the art results for these two cases. The difference between the domains is significant in these cases, and our method was not trained to optimally for these cases. The proposed method has obtained better results in all other cases, and even in these two cases, our results are competitive. 

For the Office-Home dataset, the results are reported in Table~\ref{tbl:home_office}.  For this more challenging dataset, we have achieved state-of-art performance. It is noteworthy that the proposed model provides the classification accuracy that is substantially better on this Office-Home dataset which is harder dataset for domain adaptation problem obtaining on average an improvement of 7.4\% and 7.3\% over the state of the art methods using CADA-P and CADA-A respectively.

The results on the ImageCLEF are reported in Table~\ref{tbl:imageclef}. Both CADA-P and CADA-A outperform the other state of the art models for all the transfer tasks except I$\rightarrow$C, with 0.8\% and 0.7\% improvement on average over the state of the art methods respectively.
The room for improvement is smaller than the Office-Home Dataset, as ImageCLEF only have 12 classes and datasets in each domain, and the class category is equal, making it much easier for domain adaptation.

\section{ Ablation Study }
\vspace{-0.1cm}
We investigate the Bayesian model with and without attention for both Alexnet and ResNet model on the office-31 dataset. From Table~\ref{tbl:alex_office_table} and ~\ref{tbl:res_office_table}, it is clear that the Bayesian model without attention (CADA-W) performs significantly better than the most of the other previous models, as predicting uncertainty for discriminator reduces negative transfer, by neglecting the regions which contain data uncertainty. Table~\ref{tbl:alex_office_table} and ~\ref{tbl:res_office_table} demonstrates that CADA-P (predictive certainty) performs better than CADA-A (aleatoric certainty), as predictive uncertainty includes both model and aleatoric uncertainty, providing a better estimate of certain regions for the discriminator.

\vspace{-0.2cm}
\section{Empirical Analysis}
\vspace{-0.1cm}
We further provide empirical analysis in terms of qualitative analysis of attention maps, feature visualization, discrepancy distance and statistical significance for additional insights about the performance of our method.
\subsection{Qualitative Analysis of Attention Maps}
\begin{figure}[!ht]
     \centering
      \includegraphics[scale=0.235]{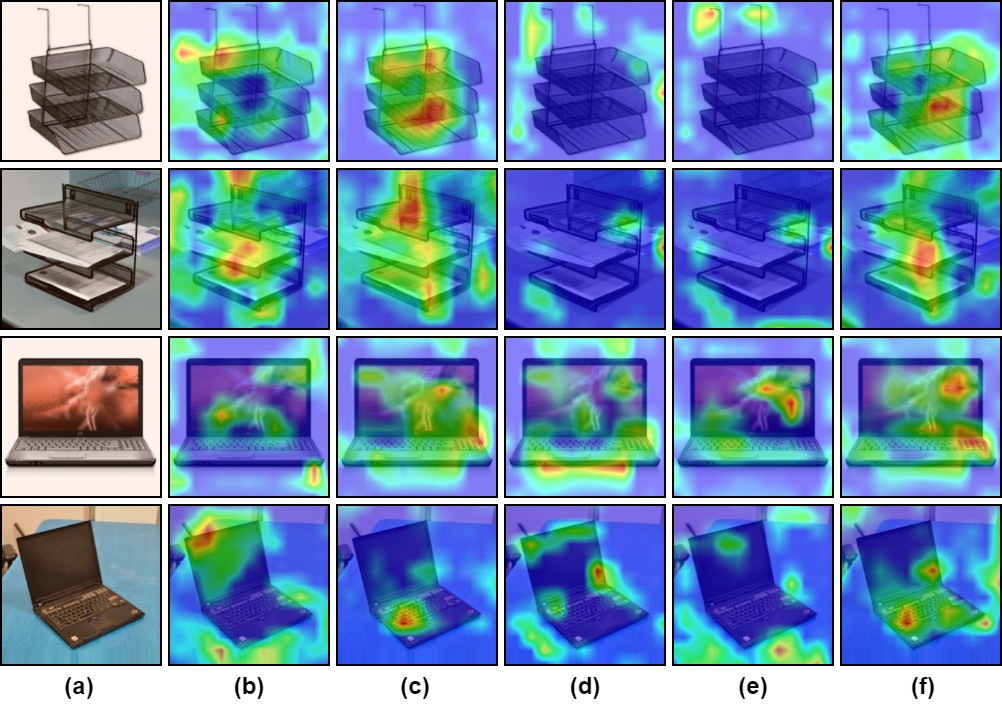}
      \caption{Attention visualization of the last convolutions layer of the proposed model CADA-P. The first and the third row shows the image from source domain (\textbf{A}) whereas the second and the fourth row shows the image from target domain  (\textbf{W}). In each row, the leftmost image (\textbf{a}) represents the original image and the rightmost image (\textbf{f}) represents the classifier's activation maps for ground truth class label at the end of the training. From left to right, the attention map of discriminator's predictive certainty is illustrated at different training stages: (\textbf{b}) 4 epochs, (\textbf{c}) 125 epochs, (\textbf{d}) 535 epochs, and (\textbf{e}) 1300 epochs. We can see as the training progress, the discriminator's certainty activation map changes from the background to the foreground, and then to the regions which can not be adapted further.}
      \label{fig:dis_train}
      \vspace{-0.6em}
 \end{figure}

\vspace{-0.1cm}
To provide the effectiveness of proposed certainty based adaptation, we provide the certainty map of the discriminator at different training stages (chosen randomly) in the Figure~\ref{fig:dis_train}. In the figure, we see that at the initial phase of the training (after 4 epochs), the discriminator discriminates the source and target domains just by some random location. As the training progress, the discriminator learns the domain by attending the background of the images (In A$\rightarrow$W, domains are mostly dissimilar in the background). After some more training, the background is adapted (after 125 epochs), and now the discriminator attends to foreground part of the image to differentiate the domains (after 535 epochs). However, the foreground varies a lot across all the images. Hence discriminator is highly certain in the class object regions. Now with further training of the model, these class object regions are also adapted (after 1300 epochs). The remaining regions of the image cannot be further adapted because there is data uncertainty.  At the end of the training, the discriminator will be highly uncertain regarding the domain, and the attention weight on the regions which are not adaptable will have low weights as we are using the certainty of the discriminator as the measure for the weights. Note that at the time of inference, we do not use the attention weights obtained by certainty for aiding classifier. These are used only at the time of training. The results show that these attention weights based training aids the classifier to better generalize to the target domain. We have provided more visualization examples in the supplementary material for further justification.
\subsection{Feature Visualization}
\vspace{-0.3em}
Adaptability of the target to source features can be visualized using the t-SNE embeddings of images feature. We follow similar setting in \cite{tzeng_arxiv2014,ganin_ICML2015,pei_arxiv2018} and plot t-SNE embeddings of the dataset in the Figure~\ref{fig:tsne}. We can observe that the proposed model correctly aligns the source and target domain images with exactly 31 clusters which are equal to the number of class labels with clear boundaries.
\subsection{Discrepancy Distance}
\vspace{-0.4em}
 $\mathcal{A}$-distance is a measure of cross domain discrepancy\cite{ben_ML2010}, which, together with the source risk, will bound the target risk. The proxy $\mathcal{A}$-distance is defined as $d_{\mathcal{A}} = 2 (1 - 2\epsilon)$, where $\epsilon$ is the generalization error of a classifier(e.g. kernel SVM) trained on the binary task of discriminating source and target. Figure \ref{fig:tsne} (e) shows $d_{\mathcal{A}}$ on tasks
A $\rightarrow $W with features of ResNet\cite{he2016deep}, GRL\cite{ganin_ICML2015}, and our model. We observe that $d_{\mathcal{A}}$ using our model features is much smaller than  $d_{\mathcal{A}}$ using ResNet and GRL features, which suggests that our features can reduce the cross-domain gap more effectively. 
\subsection{Statistical Significance Test}
\vspace{-0.4em}
We analyzed statistical significance~\cite{demvsar_JMLR2006} for variants of proposed method against GRL~\cite{ganin_ICML2015}. The Critical Difference (CD) for Nemenyi test depends upon the given confidence level (which is 0.05 in our case) for average ranks and number of test datasets. If the difference in the rank of the two methods lies within CD, then they are not significantly different. Figure~\ref{fig:ssa1} visualizes the post hoc analysis using the CD diagram for A$\rightarrow$W dataset. From the figures, it is clear that our models are significantly different from GRL\cite{ganin_ICML2015}.

\vspace{-0.2cm}
\section{Conclusion}
In this paper, we propose the use of certainty estimates of the discriminator to aid the generalization of the classifier by increasing the attention of the classifier on these regions. As it can be observed through our results, the attention maps obtained through certainty agree well with the classifier certainty for true labels and this aids in the generalization of the classifier for the target domain as well. The proposed method is thoroughly evaluated by comparison with state of the art methods and shows improved performance over all the other methods. Further, the analysis is provided in terms of statistical significance tests, discrepancy distance, and visualizations for better insight about the proposed method. The proposed method shows a new direction of using probabilistic measures for domain adaptation, and in the future, we aim to further explore this approach.

\vspace{1em}

\textbf{Acknowledgment}: We acknowledge travel support from Microsoft Research India and Google Research India. We also acknowledge resource support from Delta Lab, IIT Kanpur. Vinod Kurmi acknowledges
support from TCS Research Scholarship Program.
\newpage
{\small
\bibliographystyle{ieee_fullname}
\bibliography{egpaper_final}
}

\end{document}